%% file: main.tex
\documentclass{article}

\usepackage{microtype}
\usepackage{graphicx}
\usepackage{subfigure}
\usepackage{booktabs}
\usepackage{caption}
\usepackage{amssymb}
\usepackage{amsmath}
\usepackage{enumitem}
\usepackage{hyperref}
\usepackage{url}
\usepackage{graphicx}
\usepackage{adjustbox}
\usepackage{bm}
\usepackage[page]{appendix}
\usepackage{tabularx}
\graphicspath{{./images/}{.}}
\makeatletter
\def\input@path{{./images/}{./}}
\makeatother

\usepackage{hyperref}

\usepackage[accepted]{icml2021}

\icmltitlerunning{Graph convolutions that can finally model local structure}

\begin{document}

\twocolumn[
\icmltitle{Graph convolutions that can finally model local structure}



\icmlsetsymbol{equal}{*}

\begin{icmlauthorlist}
\icmlauthor{Rémy Brossard}{to}
\icmlauthor{Oriel Frigo}{to}
\icmlauthor{David Dehaene}{to}

\end{icmlauthorlist}

\icmlaffiliation{to}{AnotherBrain, Paris, France}

\icmlcorrespondingauthor{Rémy Brossard}{remy@anotherbrain.ai}

\icmlkeywords{Machine Learning, Autoregressive model, Graph neural network, Molecule Property Predicition, Open Graph Benchmark}

\vskip 0.3in
]

\begin{abstract}
    Despite quick progress in the last few years, recent studies have shown that modern graph neural networks can still fail at simple tasks, such as detecting small cycles \citep{loukas2019graph, chen2020can}. This suggests that current networks fail to catch information about the local structure, which is problematic if the downstream task heavily relies on graph substructure analysis, as in the context of chemistry. We propose a straightforward correction to the now standard GIN convolution \citep{Xu2018} that enables the network to detect small cycles with nearly no cost in computation time and number of parameters. Tested on real life molecule property datasets, our model consistently improves performance on large multi-tasked datasets over all baselines, both globally and on a per-task setting.
\end{abstract}

\section{Introduction}
Graph learning encompass a wide and particularly diverse set of problems. A recent surge of interest in the deep learning community and the publication of many graph datasets already lead to many successful applications, opening the way to exciting new perspectives. Graphs can naturally model the data involved in many applications, such as visual reasoning or semantic segmentation, community detection and social network predictions, recommender systems, traffic predictions, and much more\citep{zhou2018graph, wu2020comprehensive}. Among those problems, supervised whole graph property prediction is an important topic, essentially applied to molecular property prediction, which will be the center of interest of this work. Data-based approaches, as opposed to more traditional simulation-based methods,  present a great potential as the required time for a prediction could be reduced by many orders of magnitude while increasing the predictive accuracy\citep{gilmer2017neural, wu2018moleculenet, Fey2020}. Graph neural networks have been proven to be particularly effective in this task \citep{Hu2020open, wu2018moleculenet}, exceeding other approaches, both based on other data representations or on manual feature engineering. Note that the computation time is as important an issue as the performance, as modern pipelines in pharmaceutical research often rely on scanning tens of thousands of candidate molecules \emph{in silico}, before selecting only a handful for experimental tests. 

Beside these promising perspectives, molecular property prediction still presents many difficulties. A first one is owed to the cost of producing experimental measurements of biochemical properties of compounds. Thus, despite the recent revolution of high throughput screening, public datasets are often rather small. Either they contain only a few thousands or even hundreds of molecules, or they are heavily multi-tasked and scarcely labelled, so that even apparently large datasets can contain only a few handful of positively labelled molecules. Thus, overfitting is a major problem in the context of molecule property prediction.

A second problem is that the most common framework, namely message passing neural network (MPNN) \citep{Xu2018}, is not an universal approximator of functions on graphs. More precisely, it is equivalent to the first order Weisfeiler Leman kernel (1-WL). Although such network could in principle have a strong discriminative power with enough layers, it was shown that practical networks can have trouble solving even basic structure related tasks, such as detecting small cycles\citep{loukas2019graph, chen2020can}. This seems like a serious problem, as biochemistry is mainly based on the analysis of local graph structures, such as aromatic groups. Multiple attempts to solve this apparent lack of expressiveness have been proposed. A first approach is to develop methods to train larger and deeper networks\citep{li2020deepergcn}, which can already lead to a considerable boost in performance. A few works are fundamentally diverging from the standard convolution based framework, such as invariant graph networks \citep{maron2019provably, maron2018invariant} or relational pooling \citep{murphy2019relational, chen2020can}. Despite being provably more powerful, the amount of calculation required by these methods is still prohibitively large, and they usually do not outperform more traditional methods. Another approach is to extend the current framework, to make it equivalent to higher order Weisfeiler Leman kernels \citep{morris2019weisfeiler}, or to increase the receptive field of each convolution \citep{flam2020neural, nikolentzos2020k, abu2019mixhop}. However, these approaches make considerable changes to the architecture of the network and require much more parameters than the basic approach. Finally, based on the observation that the issue arises from the presence of cycles in the input graph, it was proposed to perform coupled convolution on both the input graph and a coarser version of it from which all cycles have been removed\citep{Fey2020}. Despite interesting results, this approach gives more information to the network and considerably changes the architecture.

In this work, we propose an alternative method to tackle this issue in the simplest possible fashion. We argue that increasing the effective receptive field of \emph{each} convolution step should be enough to solve the problem when only small cycles are involved, which is mostly the case in the context of chemistry. However, the \emph{global} receptive field should not be increased, as the additional information could be ill-used. We propose a variant of the standard GIN convolution to do so with a minimal cost in term of computation time and number of parameters. We show that this approach boosts performance on larger datasets. In particular, at the time this article is written, our model ranks first on the Open Graph Database leaderboard for the PCBA dataset, which is one of the largest benchmark for molecular bio-activity prediction. More generally, our results on several standard datasets show that our method is well adapted to multi-tasked settings, increasing the performance both on average and on isolated tasks.

\section{Method}
\subsection{Graph neural network}

In graph classification or regression tasks, a network operates on a labelled graph $G(\mathcal{V}, \mathcal{E})$ where $\mathcal{V} = 1 .. N$ is the set of nodes and $\mathcal{E}\subset \mathcal{V}\times\mathcal{V}$ is the set of \emph{undirected} edges, $x_v \in \mathbb{R}^{d_n}$ is the label of the node $v\in \mathcal{V}$ and $e_{ij} \in \mathbb{R}^{d_e}$ is the label of the edge $(i, j) \in \mathcal{E}$.

A very general framework to perform graph representation learning is message passing neural network (MPNN). The network consists in L message passing layers, also called graph convolutions, defined as:

\begin{align}
\begin{split}
    m^{(l)}_{ij} &= MESSAGE \left(h^{(l-1)}_i, h^{(l-1)}_j, e_{ij} \right) \\
    a^{(l)}_{i} &= AGGREGATE \left( \left\{ m^{(l)}_{ij}\right\}_{j \in \mathcal{N}_1(i)}\right) \\
    h^{(l)}_i &= UPDATE\left(h^{(l-1)}_i, a^{(l)}_i\right)
\end{split}
\label{mp}
\end{align}
where $h_i^{(l)}$ is the embedding of the node $i$ after the l-th layer. MESSAGE and UPDATE are learned functions. The AGGREGATE function acts on a multiset, i.e. a set with repetitions, defined by the neighbours of a node $i$, $\mathcal N_1(i)=\left\{j / (i, j) \in \mathcal{E} \right\}$. This function has to accept multiple sizes of set and has to be node permutation invariant so that the network does not predict a different result when the node indexing is changed. In most cases, sum, mean or max operator is used. We set $h_i^{(0)}=x_i$. After L message passing layers, the resulting node embeddings $h_i^{(L)}$ are then globally pooled to produce a fixed size vector representation, usually using once again a sum, mean or max operator. The resulting representation lies in $\mathbb{R}^n$ and is then used as input for downstream tasks.

Each layer should be able to aggregate the multiset of its neighbours in each node without loosing information. Thus, in order to discriminate different graphs, a requirement is that the operation is injective on multisets, or at least on a large finite ensemble of multisets. \citep{Xu2018} have proposed graph isomorphism network(GIN), an efficient implementation of message passing that enables this property, making it one of the most expressive convolution. In order to add edge information in the convolution, GINE a slight variant of GIN has been proposed by \cite{Hu2020} :

\begin{align}
\begin{split}
    \text{GINE :}a^{(l)}_i &= \sum_{j \in \mathcal{N}_1(i)} \sigma\left(h^{(l-1)}_j + E(e_{ij})\right) \\
    h^{(l)}_i &= MLP\left((1+\epsilon)h^{(l-1)}_i + a^{(l)}_i\right)
\end{split}
\end{align}
where $MLP$ is a standard multi-layer perceptron, $\epsilon$ is a learned parameter, $\sigma$ is a nonlinearity such as ReLU and $E(e_{ij})$ is an embedding of the label of the edge $(i, j)$.

In addition to convolutions, \cite{li2017learning} proposed the introduction of a virtual node, initially in order to allow the network to grasp long range dependencies between the nodes. A new node, so-called the virtual node,  is added to the graph and connected to every other node. The same convolutions are applied on this node, although with specific weights, so that information about the whole graph can be used at every step. Thus if the main convolution operator is GIN, the virtual node aggregation step is written as:

\begin{align}
\begin{split}
    \text{Virtual Node :} H^{(l)} &= MLP((1+\epsilon)H^{(l-1)} + \sum_{i \in \mathcal{V}} \widehat{h}^{(l)}_j) \\
    h^{(l)}_i &= \widehat h^{(l)}_i + H^{(l)}
\end{split}
\end{align}
where $H$ is the embedding of the virtual node and $\widehat h$ refers to the node embedding before the virtual node step. 

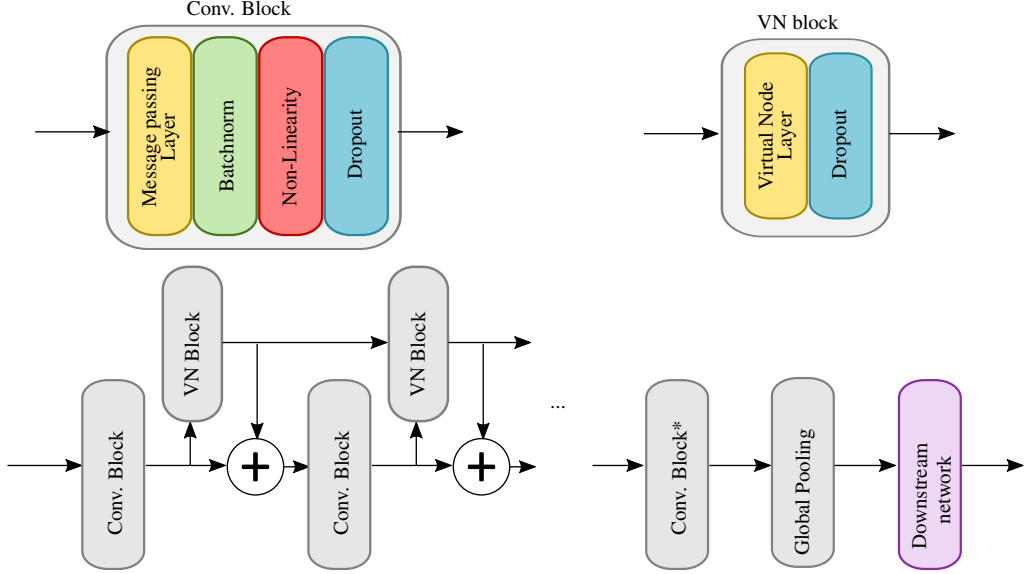
\begin{figure*}
    \centering
    \resizebox{0.8\linewidth}{!}{
    \def\svgwidth{\linewidth}
    \input{network.pdf_tex}
    }
    \caption{General structure of the network used in this work. Note that the Virtual node blocks are optional. Also note that the nonlinearity is usually omitted in the last Conv Block.}
    \label{fig:network}
\end{figure*}

The structure of such network is summed up in figure \ref{fig:network} where we added nonlinear activations and standard regularisation layers. Hyperparameters and details for the different studied tasks can be found in appendix.

\section{Increasing the kernel size for small cycle detection}

\begin{figure*}[htb]
    \centering
    \def\svgwidth{.8\linewidth}
    \input{sketch.pdf_tex}
    \caption{Illustration of the standard and the GINE+ convolutions on two different example inputs, shown in the top part of the figure. The first line in each case corresponds to the information received by the node with index $0$ during the second convolution, with radius $2$ in the case of GINE+. The second line corresponds to possible neighbourhood of node $0$ given the information it will contain after this second convolution. The distance between the node transmitting the information and node $0$ is recalled below. We can observe that the information obtained via standard convolution is not enough to infer a unique neighbourhood. Adding more convolution layers would actually not help in that case (see appendix). Note that node 0 does not have access to information from a distance greater than $2$ in this example.}
    \label{fig:sketch}
\end{figure*}
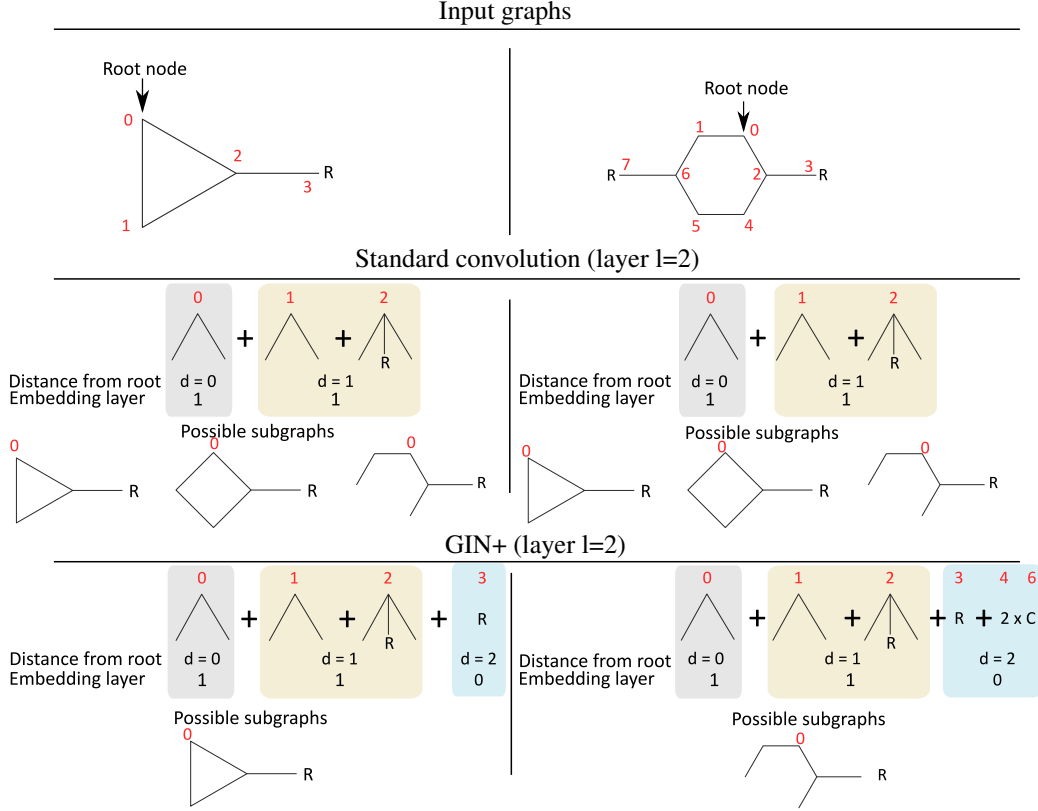

Without virtual node, convolution based networks are at most as discriminative as the Weisfeiler-Lehman test\citep{loukas2019graph}. More practically, this means that the information contained in the embedding of a node at layer $l$ can represent at best a breadth first search (BFS) tree of depth $l$ with repetition, rooted at this node. Importantly, the nodes are anonymous, so the repeated nodes between different layers are embedded as different ones. Hence, if cycles are present in the input graph, it is always possible to propose a different graph with the same embedding for a given node, as the example in figure \ref{fig:sketch} illustrates. As a result, cycle detection from the embedding of a node is fundamentally impossible. Thus, as there is no local information about the real structure of the graph, it can only be inferred after a global aggregation step where the BFS trees from each node must be assembled together. This step can become  very complex or impossible as the graph size augments, partly due to the fact that global pooling is performed with very simple functions, such as sum or mean.

In order to circumvent this problem, there are three routes. First, one could propose more complex aggregation methods, e.g. adding attention or using more complex set invariant functions. Second, one could perform the global aggregation at multiple levels of granularity to ease the task. This ultimately leads to the virtual node update. Finally, one can design a new type of convolution with a better ability to detect cycles. 

Because we believe that a method relying on global aggregation cannot be scalable, we focus on the third solution. We propose to increase the receptive field of \emph{each} convolution layer, in the simplest possible fashion. Slightly modifying the original GINE convolution, each node aggregates a wider neighbourhood, encompassing cycles if there are some. This plays a role which is the local equivalent of the global pooling layer. Hence, any cycle smaller than twice the radius of such convolution can be detected. A na\"ive implementation of this idea is:

\begin{align}
\begin{split}
    \text{NaiveGINE+ :}a^{(l, 0)}_i &= h^{(l-1)}_i \\
    a^{(l, 1)}_i &= \sum_{j \in \mathcal{N}_1(i)} \sigma(h^{(l-1)}_j + E(e_{ij}) \\
    a^{(l, k)}_i &= \sum_{j \in \mathcal{N}_k(i)} \sigma(h^{(l-1)}_j) \\
    h^{(l)}_i &= MLP\left(\sum_{k=0}^K (1+\epsilon_k)a^{(l, k)}_i\right)
\end{split}
\end{align}
Where K is the convolution radius and $\mathcal{N}_k(i)$ is the set of all nodes at distance $k$ of $i$. $\epsilon$ can be either a scalar, as in standard GIN, or a vector itself. We found it made almost no difference and used vector epsilon in this work. With $K=1$, this convolution is almost equivalent to GINE.

In principle, this formulation can enable the network to detect structures more efficiently at the local level. However, after L convolutions, the receptive field of a node is $K\times L$ so that each node has access to much more information than with standard convolution. This could be detrimental, as this information could be ill-used to build specific discriminative features, leading to overfitting instead of more meaningful ones.

Aiming at allowing local structural reasoning without adding unnecessary information, we alter the previous formulation as follows :

\begin{align}
\begin{split}
    \text{GINE+ :}a^{(l, 0)}_i &= h^{(l-1)}_i \\
    a^{(l, 1)}_i &= \sum_{j \in \mathcal{N}_1(i)} \sigma(h^{(l-1)}_j + E(e_{ij}) \\
    a^{(l, k)}_i &= \sum_{j \in \mathcal{N}_k(i)} \sigma(h^{(l-k)}_j) \\
    h^{(l)}_i &= MLP\left(\sum_{k=0}^K (1+\epsilon_k)a^{(l, k)}_i\right)
\end{split}
\end{align}

The only difference resides in the definition of $a^{(l, k)}_i$, which now takes its input in the layer $l-k$ instead of $l-1$. As a result, at layer $l$, a node will receive information from another at distance $k$, which has information about other nodes at distance at most $l-k$ from itself. Thus, the receptive distance at layer $l$ is still only $l$. Note that this amounts to signaling \emph{only} redundancies at each depth of the BFS. The behaviour of GINE+ is illustrated in figure \ref{fig:sketch}.

Notice that the new form of this convolution requires only a handful of additional parameters. Indeed, the only additions are the vector $\epsilon_k$, which contain a small number of parameters compared to the matrices in the convolution. Moreover, most of the calculations have been performed in previous layers, so that only a few supplementary additions must be executed compared to GINE.

\section{Related work}
\begin{table*}[t]
\caption{\label{tab:datasets}Main characteristics of the datasets used in this study. The first column is the total number of compounds in the dataset. The second is the number of independent classification tasks. The third is the average size of molecules in the dataset. The fourth is the proportion of data labelled as missing. The fifth is the amount of positive classes among all non missing data.  The sixth is the total number of positive classes in the dataset and the last is the total number of molecules which are positive in at least one task in the dataset.}
\vskip 0.15in
\centering
\begin{tabularx}{\linewidth}{@{}lXXXXXXX@{}}%
\toprule
Dataset & Size & Tasks & Avg. Size & Missing & Positive ratio & Positives& Pos. Molecules\\
\midrule
MUV & 93087 & 17 & 24.2 & 84 \% & 0.19 \%& 489&471\\
Tox21 &7831&12& 18.6&17\% &7.52\%& 5862&2872\\
ToxCast &8576&617& 18.8 &71\% & 8.25\%& 126651&6186\\
HIV &41127&1& 25.5&0\% &3.5\%& 1443&1443\\
PCBA &437929&128& 26&39\% &1.39\%& 472637&184126\\
\bottomrule
\end{tabularx}
\end{table*}
\subsection*{Graph convolutions}

Neural networks acting directly on graphs were described in \cite{Kipf2017} where a Graph Convolutional Network (GCN) performs a layer-wise propagation rule in the form of linear graph convolutions followed by a point-wise nonlinearity. The GCN convolution can be seen as a computationally simple first-order approximation of spectral graph convolution \citep{Bruna2013, Defferrard2016}.

While GCN is able to learn graph representations encoding structure and node features, some drawbacks have been pointed by followup works. In \cite{Xu2018} it is shown that GCNs are limited in its capability to distinguish a number of simple graph structures, and they proposed the Graph Isomorphism Network (GIN), a variant convolution which is claimed to have discriminative power equivalent to the Weisfeiler-Lehman graph isomorphism test. 

In \cite{Hu2020} it is proposed GINE, a slight modification from GIN where edge features are incorporated into the GIN convolution.

\subsection*{Substructure detection}

Even with the increased expressiveness of GIN and GINE graph convolution, its inability to properly detect graph cycles remains a major limitation, in particular when dealing with molecular graphs such as found in organic chemistry. This problem has been investigated by many recent works  \citep{chen2020can, nikolentzos2020k, abu2019mixhop, loukas2019graph, Fey2020}. In the context of molecular property prediction, we highlight the work of \cite{Fey2020} who proposed to perform message passing between the original graph and a coarser graph containing no cycles. The authors rely on fixed rules similar to the junction tree approach \citep{Jin2018} to annotate graph cycles and represent them into specific nodes.


\subsection*{Other improvements on graph neural networks}
Learning features directly on the molecular graph instead of handcrafting them or using other representation is an idea that emerged a few years ago \citep{duvenaud2015convolutional, gilmer2017neural, kong2020flag}. Many additions and modifications of the original formulation have been proposed over the years, which are complementary to our line of work. Although we did not explore this route, our method can in principle be combined with any of those other improvements. For example, regularisation or global pooling functions which may be more suited to small graphs have been proposed \citep{cai2020graphnorm, kong2020flag}, and \cite{Hu2020} have shown that pre-training strategies relying on a form of data augmentation could be beneficial.

\section{Experiments}

\begin{table*}[htb]
\caption{\label{tab:results}Test performance of the different models and datasets. (Mean $\pm$ std). Apart from our results, performances for the HIV and PCBA datasets are taken directly from the OGB leaderboard \cite{Hu2020open}. The results for DeeperGCN are obtained from the released code\protect\footnotemark. The result of HIMP on Tox21 is taken directly from the original paper and we used the released code to obtain results on MUV and ToxCast. All other performances are obtained from our implementation, with identical network structures.}
\vskip 0.15in
\begin{tabularx}{\linewidth}{@{}lXXXXX@{}}
\toprule
& MUV & Tox21 & ToxCast & HIV & PCBA\\
\hline
GCN & $0.115 \pm 0.022$ & $0.840\pm 0.004$ &  $0.735 \pm 0.002$&$0.761 \pm 0.010$ & $0.202 \pm 0.002$ \\
GCN w/ VN & $0.105 \pm 0.019$ & $0.859\pm 0.005$ & $0.743\pm 0.003$ &$0.760 \pm 0.012$ & $0.242 \pm 0.003$ \\
GINE & $0.091 \pm 0.033$ & $0.850 \pm 0.009$ & $0.741\pm 0.004 $ &$0.756 \pm 0.014$ & $0.227 \pm 0.003$\\
GINE w/ VN & $0.107 \pm 0.044$ & $0.872 \pm 0.003 $& $\mathbf{0.749\pm 0.002}$ &$0.771 \pm 0.015$ &$0.270 \pm 0.002$ \\
DeeperGCN  & $\mathbf{0.149\pm 0.023}$ & $0.831\pm 0.004$ & $0.714\pm 0.005$ & $0.786 \pm 0.012$ &$0.278 \pm 0.004$ \\
HIMP  & $0.114 \pm 0.041$  & $\bm{0.874 \pm 0.005}$ & $0.721 \pm 0.004$ & $\mathbf{0.788 \pm 0.008}$ &  $0.274 \pm 0.002 $ \\
\midrule
NaiveGINe+ K=3 w/ VN &$0.069\pm 0.0146$&$0.870\pm 0.004$&$0.737\pm 0.007$&$0.758\pm 0.013$&$0.279\pm 0.002$\\
GINe+ K=1 w/ VN & $0.062 \pm 0.025$ & $0.864\pm 0.006$ & $0.742\pm 0.002$ &$0.757 \pm 0.019$ &$0.274 \pm 0.002$ \\
GINE+ K=2 w/ VN & $0.067 \pm 0.020$ & $0.867\pm 0.003$ & $0.741\pm 0.006$ &$0.761 \pm 0.010 $&$0.286 \pm 0.003 $\\
GINE+ K=3 w/ VN & $0.106 \pm 0.026 $ & $0.867\pm 0.004$ & $\mathbf{0.749\pm 0.008}$   &$0.766 \pm 0.014$ & $\mathbf{0.292 \pm 0.002}$\\
\bottomrule
\end{tabularx}
\bigskip
\caption{\label{tab:combined results}Test performance of GINE+ compared to GINE on datasets augmented by PCBA. (Mean $\pm$ std)}
\vskip 0.15in
\begin{tabularx}{\linewidth}{@{}lXXXX@{}}
\toprule
&MUV+PCBA&Tox21+PCBA&ToxCast+PCBA & HIV+PCBA\\
\midrule
GINe w/ VN&$0.1174\pm0.0196$&$0.8810 \pm 0.0.0045$& $0.7757\pm0.0028$&$0.7875 \pm 0.0099$ \\
GINe+ K=3 w/ VN &$\mathbf{0.1194\pm0.0122}$&$\mathbf{0.8828\pm0.0043}$ &$\mathbf{0.7849\pm0.0029}$&$\mathbf{0.7919 \pm 0.0092}$ \\
\bottomrule
\end{tabularx}
\end{table*}
\footnotetext{Same arguments than for the \emph{ogbg-molhiv} dataset as given in the documentation, except that we set the number of layers to 3 and the number of hidden dimensions to 100 for comparability.}

\subsection{Dataset}
\label{section:datasets}

We performed experiments on five molecule classification datasets, from MoleculeNet \cite{wu2018moleculenet} and Open Graph Benchmark (OGB) \cite{Hu2020open} collections, which cover a wide range of tasks:

\begin{itemize}[itemsep=0ex]
    \item MUV is a dataset designed from PubChem bioactivity data specially designed for virtual screening methods benchmark.
    \item Tox21 and ToxCast contain molecules assayed against diverse toxicity-related targets, based on in vitro high-throughput screening.
    \item HIV is a single classification task dataset corresponding to the experimentally measured ability of a compound to inhibit the HIV virus replication.
    \item \nopagebreak{PCBA is a subset of PubChem BioAssay consisting of measured bio-activities of compounds.}
\end{itemize}

The characteristics of all datasets are summed up in table \ref{tab:datasets}. We noted that, despite sometimes deceptively large sizes, molecule datasets are extremely skewed. In particular, most datasets contain only a few thousand molecules of interest, in the sense that they are positively classified in at least one task. These datasets can be overfitted easily: if those few molecules are memorized, the network only has to be able to differentiate distinct molecules. This is a task for which graph neural networks are particularly well fitted. PCBA is the only dataset that does not suffer from this lack of positive samples: Being both large and heavily multi-tasked, it contains nearly two orders of magnitude more molecules of interest.

Different kind of splitting methods are typically used. Random splitting is simple, but can produce misleading results as there are a lot of small variations of the same sample in most molecule datasets. Scaffold splitting, where samples are separated in based on their general structure, is often used to perform experiments closer to the reality. In this work, we use the scaffold split provided with the Open Graph Benchmark framework for the MUV, HIV and PCBA datasets and random split for the Tox21 and ToxCast datasets.

For comparability with litterature, we use PRC-AUC as the metric for the two most skewed datasets, MUV and PCBA, while we use ROC-AUC on the others.

Finally, we augment the smallest datasets with PCBA, making them a supplementary task of the larger PCBA dataset. This is done to prevent overfitting by increasing the number of samples. Validation and test are still performed only on the smaller sets. These datasets are used to study the performance on \emph{only one} task in a multitasked setting. 

\subsection{Training details}
We train a classifier network by minimizing cross-entropy loss with the target properties. Details about the architecture and hyperparameters are given in annex.

All experiments are run five times on a GeForce GTX 1080 Ti GPU before measuring their average performance and standard deviation. The longest experiments are those on the PCBA dataset or datasets augmented by PCBA, which take approximately 6 hours.

The code is available online, the URL will be announced after the anonymous review process is finished.
\subsection{Results}

First, we evaluate the performance of our model against baselines on all datasets. The results are displayed in table \ref{tab:results}. Our model ranks first on the large PCBA dataset, by a clear margin, making it the leading model for PCBA in the OGB leaderboard when this paper was written. However, on the four smaller datasets, the model does not outperform state-of-the-art models. In particular, it performs slightly lower than GINE with virtual node, which shares most similarities with our model. We believe this is due to the fact that the smaller datasets are easier to overfit, so that the addition in expressive power in our model was ill-used to memorize the dataset instead of learning useful structural features.

In order to limit the ability of the model to overfit, we evaluate the performance of our model on small datasets combined with PCBA. As detailed in section \ref{section:datasets}, the training is performed on the training data of both datasets, but evaluation is performed only on the validation set of the non-augmented dataset. Obviously, we observe a performance boost over the training on the non-augmented datasets. Nonetheless, our model performs better on all augmented datasets compared to GINE.

Finally, we note that these improvements come at a very small price: in our experiments, the calculation time increased by 3 to 15 \% while the increment in number of parameters is only a few fractions of percent.

\section{Conclusion}

We proposed GINE+ which builds on the idea of the standard GIN convolution by increasing the radius of the convolution in a non-naive way. Without increasing the perceptive field, which could lead to undesirable effects, such as overfitting, GINE+ produce node embeddings that are more informative about its local neighbourhood. In particular, it can efficiently learn to detect small cycles. GINE+ comes at a small cost in term of computation time and a negligible one in term of number of parameters.

We apply our method to molecular property prediction. Our model consistently improves performance on large datasets, and seems to produce useful features that transfer efficiently in a multitask setting. In particular, our model ranks first on the PCBA dataset when this article is written.

Beside this result, we believe that the ability to infer a node neighbourhood from its embedding is essential in numerous graph problems, in particular in the case of graph pooling and graph generation. We leave these development to future works.

\bibliography{biblio}
\bibliographystyle{icml2021}

\newpage
\renewcommand{\appendixpagename}{Appendix}
\begin{appendices}
\section{Architecture and hyperparameters}
All networks trained have the same structure:
\begin{itemize}
    \item A node embedding layer which produce an embedding vector $x_i^{(0)}$ of dimension $H$. We use the AtomEmbedding function from the Open Graph Benchmark package.
    \item $L$ layers of convolution with batch normalization, ReLU activation and dropout layer (p=0.5) with constant output dimension $H$, eventually with radius $K$.
    \item Optionally, a virtual node with same dimension H.
    \item A global mean pooling layer
    \item A linear classifier
\end{itemize}

Each layer has its own trainable weights.

GIN type convolutions (GINE, NAIVEGINE+, GINE+, with or without virtual node) require a MLP. We use a basic implementation with both input and output dimensions set to $H$ and a hidden layer of dimension $2H$ followed by batch normalization, ReLU activation and dropout.

For the PCBA dataset, we used $L=5$ and $H=400$. For all other datasets, we used $L=3$ and $H=100$. Those value where selected by hand, for $L\in {2,3,4,5}$ and $H\in{100,200,300,400}$. The radius $K$ was varied between 1 and 5, without further improvement above 3.

The model was always trained for 100 epochs using the Adam optimizer with a learning rate of 0.001. The implementation was written using Pytorch, Pytorch-Lightning and Pytorch-Geometric libraries.

The code is implemented using PyTorch and Pytorch-Lightning.

All datasets are taken from the Open Graph Benchmark package, and eventually shuffled when a random split is used.

\section{Existence of graphs that cannot be differentiated by standard convolutions}
\label{undifferentied graphs}

From a graph $G(V, E)$ of size $N$ containing at least one cycle, we show that there exists another graph containing at least $N$ nodes with the exact same embeddings after L layers of convolution than the nodes of the original graph, for any L.

Consider the graph $G_{copy}(V', E')$, isomorphic to $G$ with $V \cap V' = \varnothing$ i.e. a copy of the first graph without redundant indexing. We denote $\phi$ the isomorphism so that $V'=f(V)$. Given a specific edge $(i, j) \in E$ and $(i', j')=(\phi(i), \phi(j)) \in E'$ the corresponding edge in  $G_{copy}$. We build a new edge set $\widehat{E} = (E' \cup E) \cup \left\{(i, j'), (i', j)\right\} / \left\{(i, j), (i', j')\right\}$. That is we replace the edge $(i, j)$ by $(i, j')$ and reciprocally. The final graph is $\widehat{G}(V \cap V', \widehat{E})$. If edge labels are present, the label of the two added edges are the same than the two removed edges.

We note $x_k^{(l)}$ and $\widehat x_k^{(l)}$ the node embedding of node $k$ in graph $G$ and $\widehat{G}$ respectively after l layers of convolution. We show recursively that $x_k^{(l)}=\widehat x_k^{(l)}=\widehat x_{\phi(k)}^{(l)}$  for any $l > 0$ and $k\in V$.

First, for any $k \in V$, since initial node embedding does not depend of the neighbourhood, $x_u^{(0)}=\widehat x_u^{(0)}=\widehat x_{\phi(u)}^{(0)}$ for any $u\in V$. 

Then, for a given $l > 0$ and $k\in V$, if $x_u^{(l-1)}=\widehat x_u^{(l-1)}=\widehat x_{\phi(u)}^{(l-1)}$ for any $u \in V$, then we consider three cases: First, if $k \ne i\text{ or }j$, the set of the embeddings of its neighbours is identical in $G$ and $\widehat{G}$. Second, if $k=i$, the set of the embeddings of its neighbours in $\widehat{G}$ is the same as in $G$ except without $x_j^{(l-1)}$ and with the addition of $\widehat{x}_j^{(l-1)}$, which are equal by construction so that the two set are still identical. An identical argument goes for the last case $k=j$. Due to the definition of message passing in equation \ref{mp}, this implies that $x_u^{(l)}=\widehat x_u^{(l)}=\widehat x_{\phi(u)}^{(l)}$ for any $u \in V$, which concludes the proof.

\end{appendices}
\end{document}

%% file: images/network.pdf_tex
\begingroup%
  \makeatletter%
  \providecommand\color[2][]{%
    \errmessage{(Inkscape) Color is used for the text in Inkscape, but the package 'color.sty' is not loaded}%
    \renewcommand\color[2][]{}%
  }%
  \providecommand\transparent[1]{%
    \errmessage{(Inkscape) Transparency is used (non-zero) for the text in Inkscape, but the package 'transparent.sty' is not loaded}%
    \renewcommand\transparent[1]{}%
  }%
  \providecommand\rotatebox[2]{#2}%
  \newcommand*\fsize{\dimexpr\f@size pt\relax}%
  \newcommand*\lineheight[1]{\fontsize{\fsize}{#1\fsize}\selectfont}%
  \ifx\svgwidth\undefined%
    \setlength{\unitlength}{573.14501665bp}%
    \ifx\svgscale\undefined%
      \relax%
    \else%
      \setlength{\unitlength}{\unitlength * \real{\svgscale}}%
    \fi%
  \else%
    \setlength{\unitlength}{\svgwidth}%
  \fi%
  \global\let\svgwidth\undefined%
  \global\let\svgscale\undefined%
  \makeatother%
  \begin{picture}(1,0.56557399)%
    \lineheight{1}%
    \setlength\tabcolsep{0pt}%
    \put(0,0){\includegraphics[width=\unitlength,page=1]{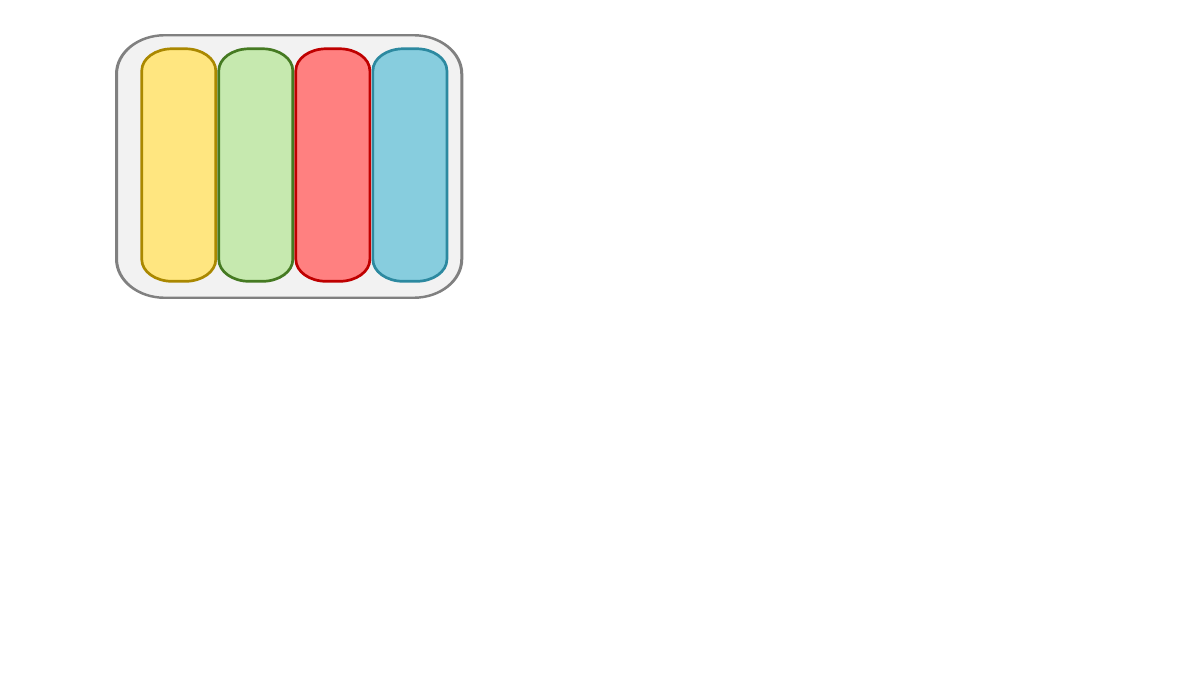}}%
    \put(0.17660991,0.54658611){\color[rgb]{0,0,0}\makebox(0,0)[lt]{\lineheight{1.25}\smash{\begin{tabular}[t]{l}Conv. Block\end{tabular}}}}%
    \put(0,0){\includegraphics[width=\unitlength,page=2]{network.pdf}}%
    \put(0.73772362,0.5328999){\color[rgb]{0,0,0}\makebox(0,0)[lt]{\lineheight{1.25}\smash{\begin{tabular}[t]{l}VN block\end{tabular}}}}%
    \put(0,0){\includegraphics[width=\unitlength,page=3]{network.pdf}}%
    \put(0.11251224,0.0346042){\color[rgb]{0,0,0}\rotatebox{89.320737}{\makebox(0,0)[lt]{\lineheight{1.25}\smash{\begin{tabular}[t]{l}Conv. Block\end{tabular}}}}}%
    \put(0,0){\includegraphics[width=\unitlength,page=4]{network.pdf}}%
    \put(0.18780532,0.17050915){\color[rgb]{0,0,0}\rotatebox{89.320737}{\makebox(0,0)[lt]{\lineheight{1.25}\smash{\begin{tabular}[t]{l}VN Block\end{tabular}}}}}%
    \put(0,0){\includegraphics[width=\unitlength,page=5]{network.pdf}}%
    \put(0.33497016,0.03460419){\color[rgb]{0,0,0}\rotatebox{89.320737}{\makebox(0,0)[lt]{\lineheight{1.25}\smash{\begin{tabular}[t]{l}Conv. Block\end{tabular}}}}}%
    \put(0,0){\includegraphics[width=\unitlength,page=6]{network.pdf}}%
    \put(0.41026408,0.17050834){\color[rgb]{0,0,0}\rotatebox{89.320737}{\makebox(0,0)[lt]{\lineheight{1.25}\smash{\begin{tabular}[t]{l}VN Block\end{tabular}}}}}%
    \put(0,0){\includegraphics[width=\unitlength,page=7]{network.pdf}}%
    \put(0.53422815,0.15985887){\color[rgb]{0,0,0}\makebox(0,0)[lt]{\lineheight{1.25}\smash{\begin{tabular}[t]{l}...\end{tabular}}}}%
    \put(0,0){\includegraphics[width=\unitlength,page=8]{network.pdf}}%
    \put(0.6673063,0.03460431){\color[rgb]{0,0,0}\rotatebox{89.320737}{\makebox(0,0)[lt]{\lineheight{1.25}\smash{\begin{tabular}[t]{l}Conv. Block*\end{tabular}}}}}%
    \put(0,0){\includegraphics[width=\unitlength,page=9]{network.pdf}}%
    \put(0.78858236,0.01433647){\color[rgb]{0,0,0}\rotatebox{89.320737}{\makebox(0,0)[lt]{\lineheight{1.25}\smash{\begin{tabular}[t]{l}Global Pooling\end{tabular}}}}}%
    \put(0,0){\includegraphics[width=\unitlength,page=10]{network.pdf}}%
    \put(0.90405173,0.0208781){\color[rgb]{0,0,0}\rotatebox{89.320737}{\makebox(0,0)[lt]{\lineheight{1.25}\smash{\begin{tabular}[t]{l}Downstream\end{tabular}}}}}%
    \put(0.92760596,0.04556235){\color[rgb]{0,0,0}\rotatebox{89.320737}{\makebox(0,0)[lt]{\lineheight{1.25}\smash{\begin{tabular}[t]{l}network\end{tabular}}}}}%
    \put(0,0){\includegraphics[width=\unitlength,page=11]{network.pdf}}%
    \put(0.14626158,0.42667058){\color[rgb]{0,0,0}\rotatebox{89.941069}{\makebox(0,0)[t]{\lineheight{0.75}\smash{\begin{tabular}[t]{c}Message passing\\Layer\end{tabular}}}}}%
    \put(0.34882679,0.38531307){\color[rgb]{0,0,0}\rotatebox{90}{\makebox(0,0)[lt]{\lineheight{1.25}\smash{\begin{tabular}[t]{l}Dropout\end{tabular}}}}}%
    \put(0.22242778,0.37239018){\color[rgb]{0,0,0}\rotatebox{90}{\makebox(0,0)[lt]{\lineheight{1.25}\smash{\begin{tabular}[t]{l}Batchnorm\end{tabular}}}}}%
    \put(0.28450019,0.35903189){\color[rgb]{0,0,0}\rotatebox{90}{\makebox(0,0)[lt]{\lineheight{1.25}\smash{\begin{tabular}[t]{l}Non-Linearity\end{tabular}}}}}%
    \put(0.7532215,0.42797071){\color[rgb]{0,0,0}\rotatebox{90}{\makebox(0,0)[t]{\lineheight{0.75}\smash{\begin{tabular}[t]{c}Virtual Node\\Layer\end{tabular}}}}}%
    \put(0.82553692,0.3855463){\color[rgb]{0,0,0}\rotatebox{90}{\makebox(0,0)[lt]{\lineheight{1.25}\smash{\begin{tabular}[t]{l}Dropout\end{tabular}}}}}%
  \end{picture}%
\endgroup%

%% file: images/sketch.pdf_tex
\begingroup%
  \makeatletter%
  \providecommand\color[2][]{%
    \errmessage{(Inkscape) Color is used for the text in Inkscape, but the package 'color.sty' is not loaded}%
    \renewcommand\color[2][]{}%
  }%
  \providecommand\transparent[1]{%
    \errmessage{(Inkscape) Transparency is used (non-zero) for the text in Inkscape, but the package 'transparent.sty' is not loaded}%
    \renewcommand\transparent[1]{}%
  }%
  \providecommand\rotatebox[2]{#2}%
  \newcommand*\fsize{\dimexpr\f@size pt\relax}%
  \newcommand*\lineheight[1]{\fontsize{\fsize}{#1\fsize}\selectfont}%
  \ifx\svgwidth\undefined%
    \setlength{\unitlength}{514.01785302bp}%
    \ifx\svgscale\undefined%
      \relax%
    \else%
      \setlength{\unitlength}{\unitlength * \real{\svgscale}}%
    \fi%
  \else%
    \setlength{\unitlength}{\svgwidth}%
  \fi%
  \global\let\svgwidth\undefined%
  \global\let\svgscale\undefined%
  \makeatother%
  \begin{picture}(1,0.78412812)%
    \lineheight{1}%
    \setlength\tabcolsep{0pt}%
    \put(0,0){\includegraphics[width=\unitlength,page=1]{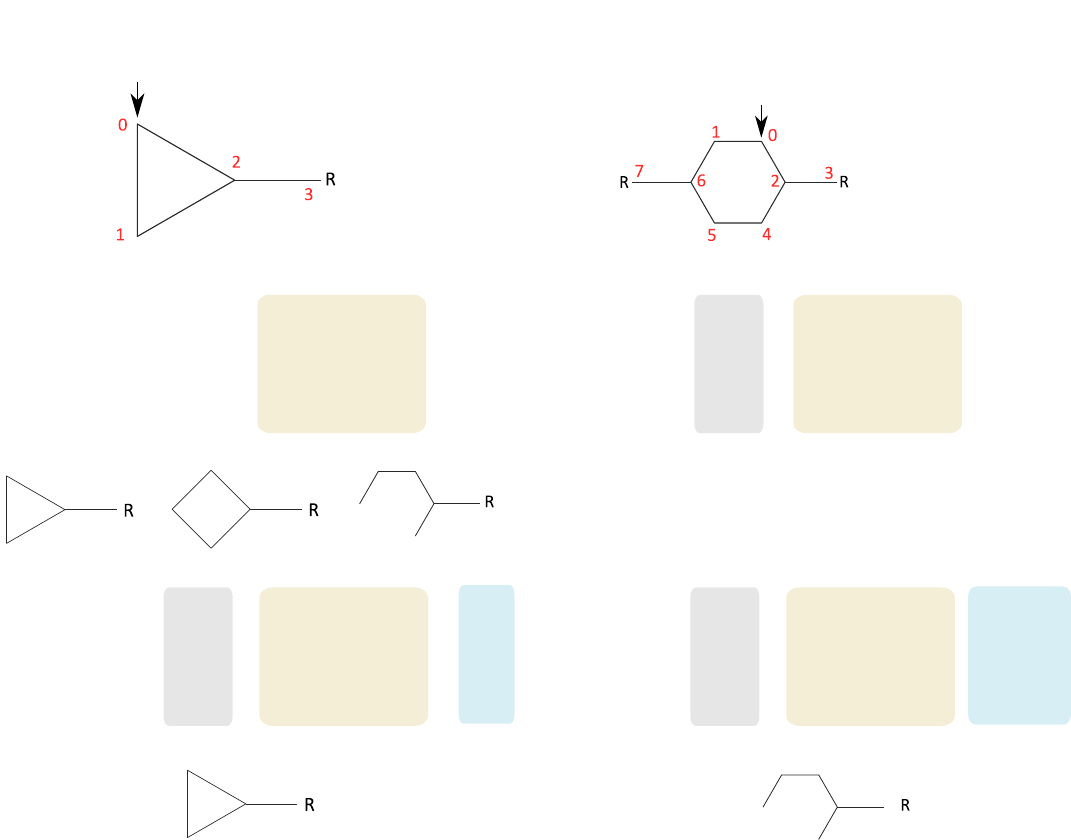}}%
    \put(0.33458112,0.52464376){\color[rgb]{0,0,0}\makebox(0,0)[lt]{\lineheight{1.25}\smash{\begin{tabular}[t]{l}Standard convolution (layer l=2)\end{tabular}}}}%
    \put(0.42157479,0.24715302){\color[rgb]{0,0,0}\makebox(0,0)[lt]{\lineheight{1.25}\smash{\begin{tabular}[t]{l}GIN+ (layer l=2)\end{tabular}}}}%
    \put(0,0){\includegraphics[width=\unitlength,page=2]{sketch.pdf}}%
    \put(0.41564177,0.76560258){\color[rgb]{0,0,0}\makebox(0,0)[lt]{\lineheight{1.25}\smash{\begin{tabular}[t]{l}Input graphs\end{tabular}}}}%
    \put(0,0){\includegraphics[width=\unitlength,page=3]{sketch.pdf}}%
  \end{picture}%
\endgroup%